# Unified Physical Threat Monitoring System Aided by Virtual Building Simulation

Zenjie Li
Milestone Systems A/S
zli@milestone.dk

Barry Norton
Milestone Systems A/S
bno@milestone.dk

*Abstract*—With increasing physical threats in recent years targeted at critical infrastructures, it is crucial to establish a reliable threat monitoring system integrating video surveillance and digital sensors based on cutting-edge technologies. A physical threat monitoring solution unifying the floorplan, cameras, and sensors for smart buildings has been set up in our study. Computer vision and deep learning models are used for video streams analysis. When a threat is detected by a rule engine based on the real-time analysis results combining with feedback from related digital sensors, an alert is sent to the Video Management System so that human operators can take further action.

A physical threat monitoring system typically needs to address complex and even destructive incidents, such as fire, which is unrealistic to simulate in real life. Restrictions imposed during the Covid-19 pandemic and privacy concerns have added to the challenges. Our study utilises the Unreal Engine to simulate some typical suspicious and intrusion scenes with photorealistic qualities in the context of a virtual building. Add-on programs are implemented to transfer the video stream from virtual PTZ cameras to the Milestone Video Management System and enable users to control those cameras from the graphic client application. Virtual sensors such as fire alarms, temperature sensors and door access controls are implemented similarly, fulfilling the same programmatic VMS interface as real-life sensors. Thanks to this simulation system's extensibility and repeatability, we have consolidated this unified physical threat monitoring system and verified its effectiveness and user-friendliness. Both the simulated Unreal scenes and the software add-ons developed during this study are highly modulated and thereby are ready for reuse in future projects in this area.

***Keywords-*** Physical threat monitoring, Unreal engine

## I. INTRODUCTION

Over the last decade, increasing physical threats targeted at critical infrastructures have been reported. Especially during the COVID-19 crisis since early 2020, related crimes targeted at health centres have severely threatened public safety[1]. At the same time, intelligent threat monitoring system has attracted significant attention from both academia and industry. With the significant performance improvement of computer vision algorithms, especially deep learning models based on Convolutional Neural Networks (CNN), intelligent surveillance systems have become more accessible in practice (see [1] for a review). In case of incidents detected via live video, events will be sent to a Video Management System (VMS) and hence to end-users. Many studies focus on video surveillance exclusively. It is, however, crucial for a reliable building threat monitoring system to integrate other digital sensors, such as door access controls, smoke/fire sensors and temperature sensors, and device location information.

We have set up a threat monitoring system unifying the floorplan, cameras, and other sensors for buildings with the widely used Milestone VMS XProtect®[2]. The goal is to effectively detect, report and handle physical threats or suspicious behaviours. When security incidents are detected, based on visual object analysis on video streams and the feedback from related digital sensors, alerts will be sent to the VMS and further to the stationary or mobile operation consoles so that human operators can act promptly.

Physical threat monitoring studies have to consider hazardous and complex scenarios, such as forceful intrusion, fire and power cut. This poses challenges for data collection and testing, especially in threat monitoring for critical infrastructure and during a pandemic. To make it possible to develop and validate our unified threat monitoring system, we have developed a virtual hospital represented in the Unreal game engine. This engine originates from the gaming industry but has been successfully applied to areas like smart city and traffic system simulation (see, e.g., [1, 2]). It is, however, to date much less used for simulating smart buildings.

In this paper, the set-up of our physical threat monitoring system is introduced in Section 2. Section 3 describes the procedures for creating a virtual smart building and integrating virtual devices into the VMS. Use cases are presented and discussed in Section 4 before conclusions are drawn in Section 5.

## II. PHYSICAL THREAT MONITORING SYSTEM

### A. System Architecture

Milestone Systems' VMS, XProtect®, is used in this study. XProtect® is a powerful VMS solution with free versions available for download. Using the Milestone Integration Platform Software Development Kit (MIP SDK[3]), one can add support for hardware devices, either physical or virtual, and add new custom software features. Our threat monitoring system includes a video analytics component, a data analytics component (rule engine), and event forwarding services integrated with the XProtect®. The system architecture is shown in Figure 1. Video analytics, built on the Milestone Video Processing Service (VPS) platform [4], receives video streams from cameras. After processing, VPS sends results as metadata back and publishes the results via Apache Kafka[5]. A rule engine subscribes to the Kafka events and makes decisions using the static data from the graph database and events generated by other related sensors. New events representing alarms are then published on Kafka. Those are forwarded to the VMS Event Server and visualised on a graphic user application, either on a desktop or mobile device, so that the end-users can acknowledge or reject the alarms.

---

[1] https://www.safecare-project.eu/?p=588.

[2] https://www.milestonesys.com/solutions/platform/video-management-software/.

[3] https://doc.developer.milestonesys.com/html/index.html.

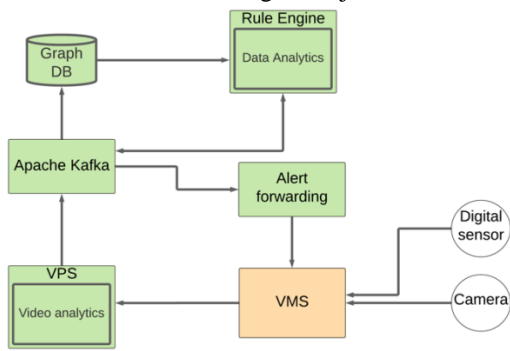

Figure 1. Architecture diagram.

The video analytics and the rule engine (data analytics) are separated into two modules. The former is often deep-learning-based, while the latter is often not, and the publicly available libraries are often very different for them. It is noteworthy that some digital sensors, such as smoke/fire, temperature and flood sensors, can trigger alarms in the VMS without video analytics and data analytics. Still, when the alarms are propagated to the user applications, related videos will also be shown.

### B. Threats Monitored Using Visual Object Detections

Physical threats, such as tailgating, crowding, loitering and weapons, are based on video analysis. As an example, the workflows for tailgating detection are elaborated here. Other threats in this category are detected based on similar principles.

Tailgating is a behaviour in which an unauthorised person follows an authorised one when passing an access-controlled gate, without identifying themselves via access card. Anti-tailgating detectors based on optical sensors are available on the market. However, without extra hardware cost, a vision-based tailgating detector can be implemented using the existing security cameras and door access controls. A simple solution has been implemented in our study, which detects tailgating using a door-facing camera mounted inside the room Figure 2(a)) and door access control events. This software monitors and tracks persons in the camera view and compares their locations relative to the door location (the white box) pre-stored on the database. When a person has been detected passing the door, a door transition event will be published on Kafka.

Visual object detection (finding each object's location and class in one video frame) and visual object tracking (assigning a unique ID to the same object in different video frames) are essential parts of tailgating detection. In our system, the state-of-the-art YOLOv3 [3] is used for object detection due to the excellent balance of speed and accuracy. An off-the-shelf implementation provided by NVIDIA Deepstream [6], based on discriminative correlation filters, is used for tracking.

The software detects a person at the door when the relative overlapping of the person bounding box (area denoted as $S_{person}$) and the static door bounding box (area denoted as $S_{door}$) is larger than a threshold $\delta$, i.e., $(S_{door} \cap S_{person})/S_{door} > \delta$. In our study, an empirical value $\delta = 0.5$ is used. When a person is not at the door anymore (Figure 2(b, c)), a door transition event is generated and published on Kafka. Since synchronised video and door access control events are used in tailgating detection, a virtual environment will make development and validation much more accessible.

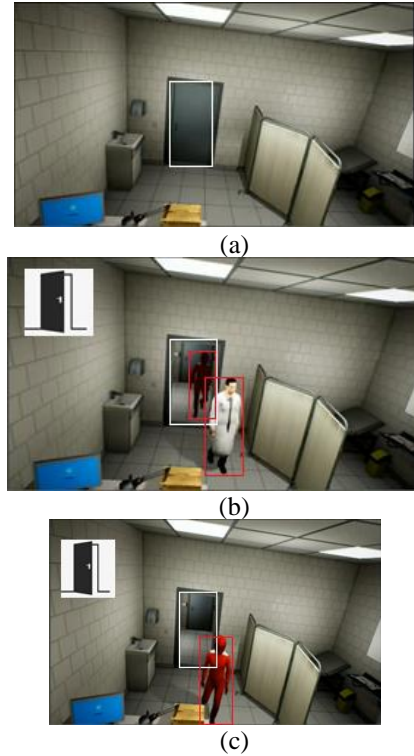

Figure 2. Camera view from inside a door. Images from the virtual hospital (see Section 3). (a) The door is closed, and the door location is pre-stored in the database (marked by the white box); (b) The door is opened, two people are detected, and one person has entered; (c) The door is opened, and the second person has entered.

---

[4] https://doc.developer.milestonesys.com/html/samples/VPService_sample.html.
[5] https://kafka.apache.org/.
[6] https://developer.nvidia.com/deepstream-sdk.

## C. Other Threat Monitoring Types

Some threats such as abnormal temperature and flood are detected solely with digital sensors. Fire can be detected both by smoke/fire sensors and surveillance cameras. Smoke/fire sensors are economical and easy to install but with high false-positive rates. The advantages of video-based are the availability of a confidence value and the possibility to monitor a large space difficult for smoke/fire sensors. Binary image classification is used for distinguishing images with alarming fire/smoke from those without. A light-weight model, ResNet-18 **[4]**, is used in our study.

Lack of data is one significant challenge in detecting hazardous incidents. The data used in our work came from two sources. The first source is videos/images downloaded from publicly available data sets (USTC fire detection research group[7] and the CAIR fire detection dataset[8]. The second source is videos simulated in-house using the Unreal Engine.

## III. VIRTUAL BUILDING

A virtual hospital including virtual cameras and other digital sensors has been created in our study to address the development and testing issues.

Virtual building establishment starts from creating a 3D building model based on some existing templates or 2D floorplan if the virtual building needs to be a faithful representation of a real-world building. The 2D-to-3D conversion can be carried out manually with software tools or automatically using 3D building generation algorithms [5]. Upon completion, the 3D building can be exported to industrial standard formats, such as *.obj* and *.fbx* files.

To be eligible for a threat monitoring study, one needs to fill a 3D virtual building model with human characters, security devices and other supporting assets. Animations are used to simulate physical threat scenarios, and the virtual building should interact with external applications and eventually with end-users. The Unreal game engine is used in our work to achieve this goal. Unreal is used because it is free for internal and free projects [9] and provides photorealistic rendering capabilities. Its visual scripting tool, BluePrint, makes it easier for beginners to create basic scenes while more experienced developers can leverage the power of C++. Unreal Engine version 4.25 has been tested in the current work.

Implementing virtual cameras, other digital sensors, and remote user interactions are elaborated in the following.

### A. Virtual Cameras

Virtual cameras in this virtual building rely on the Unreal component, *SceneCaptureComponent2D*, for image capturing. This component captures the scene from its view frustum for each frame and stores it as an image used within a Material. The technique has been widely used for representing mirrors, mini-maps, and security cameras in games.

In our implementation, the virtual camera BluePrint (*BP_VirtualCamera*) containing a *SceneCaptureComponent2D* is inherited from an Unreal Actor implemented in C++, which maintains an HTTP server. External client applications, such as web browsers or customised HTTP client applications, can receive the video stream captured by *SceneCaptureComponent2D* via the URL configured on the particular camera's HTTP server. Figure 3 shows the Unified Modeling Language class diagram for a virtual camera class.

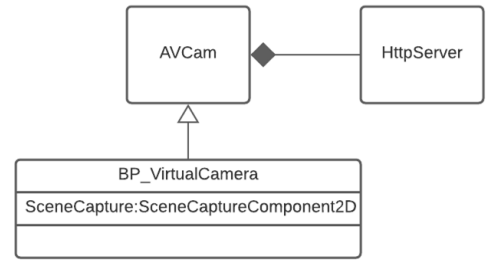

Figure 3. Unified Modeling Language class digram for virtual cameras.

For the virtual cameras to be accessible for the VMS, a virtual device driver has been developed with the *Milestone Driver Framework*[10]. The virtual device driver implements an HTTP client, which reads the images at a user-defined frame rate from the virtual camera's HTTP server. Then it pushes the images to the XProtect® Server. Moreover, by translating the pan, tilt, and zoom commands from the device driver into rotations and field of view changes, one can simulate pan-tilt-zoom (PTZ) cameras. The virtual camera appears in XProtect® as an ordinary camera.

### B. Non-video Digital Devices

When a client application, say a VMS device driver, is connected to a virtual camera in a virtual building and requests data, it will receive a continuous data stream from the camera. In contrast, non-video digital devices, such as fire alarms, door access controls and temperature sensors, does not respond to client requests. Instead, they send data in case of state change, e.g., when a fire alarm is activated or deactivated. HTTP communication is not enough for those sensors, and we need WebSocket, a full-duplex communication protocol.

An alarm manager blueprint the virtual building maintains a WebSocket server and a list of connected client IDs. When the state of a virtual device is changed, the alarm manager will forward the notification to the WebSocket client. By wrapping such a WebSocket client in a device driver or plugin, a virtual device will appear in XProtect® as an ordinary device.

---

[7] http://smoke.ustc.edu.cn/datasets.htm
[8] https://github.com/cair/Fire-Detection-Image-Dataset
[9] Unreal® Engine End User License Agreement for Creators: https://www.unrealengine.com/en-US/eula/creators

[10] https://doc.developer.milestonesys.com/html/gettingstarted/intro_driverframework.html

The C++ WebSocket server implementation in the current study uses *Beast* in the Boost library[11].

### C. User Request Handling

In ordinary games, user interactions are frequent and happen via the game UI itself. In the case of security incident simulation, user interaction is sporadic, and the user is not always beside the gaming console. Screen rendering might be turned off to save video memory consumption. Therefore, a REST Application Programming Interface (REST API [12]) has been implemented for users to control the game, such as trigger a door opening, a human character action, or a security incident.

Like a virtual camera, a request handler blueprint maintains an HTTP server and a list of actors with a standard action interface. Upon a request, the request handler will find the corresponding actor and triggers the action. For example, suppose the virtual building is running locally, and the user requests http://localhost:20001/action?id=0&name=fire&value=startFire. The fire emitter asset (a particle system in Unreal) will be spawned at the specific location *0*, and the fire will start.

## IV. USE CASES

The virtual building has been used to simulate various physical security incidents. It cannot replace a real-life environment but can significantly facilitate the development and validation of a physical threat monitoring system. This section demonstrates some representative use cases involving tailgating, fire and weapons, with some other threat types briefly covered. At the end of the section, suggestion for future work is provided.

### A. Tailgating

The scene repeatability and flexibility are crucial for developing and validating detection of incidents such as tailgating and loitering in restricted areas. Different sensors should often be synchronised, such as cameras and door access controls. While those requirements are challenging to achieve, it is accessible in a virtual world. In our simulated tailgating scene below, a doctor opens an access-controlled door and enters an operation room. An unauthorised person follows him, also enters the room and steals a medicine box. Two virtual cameras have been mounted, one in the corridor and another in the room. The door access control communicates with the VMS via a plugin as described in Section BThe video streams from the two cameras are shown in Smart Client, the XProtect® client application, in sync with the animations in Unreal. Figure 4 shows two screenshots of the Smart Client at two different time frames. In Figuire 4(a), the doctor walks towards the door, and the door state is "Locked". In Figure 4(b), the doctor enters the room, followed by an unauthorised person. When the follower also enters the room, our threat monitoring system detects a tailgating incident and sends an alarm (see also Section B). In this case, only the camera inside the room is used for detection.

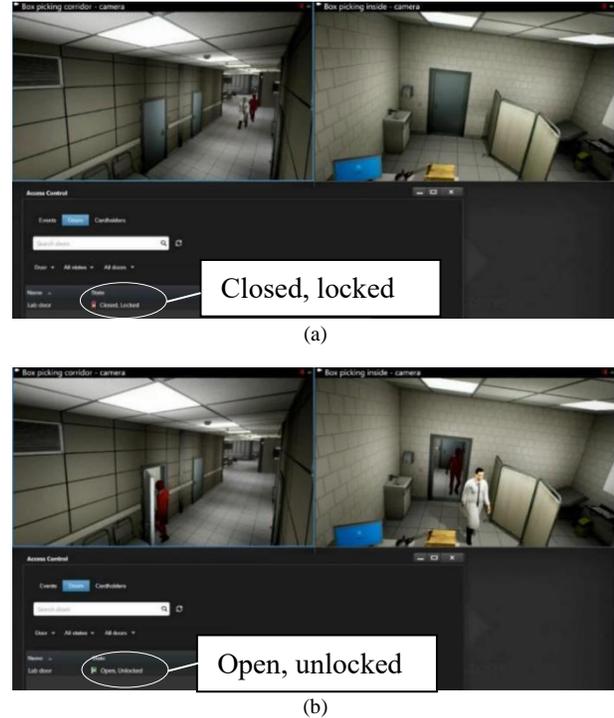

Figure 4. Two camera views of simulated tailgating scenes as visualised in Smart Client, the XProtect client application. The annotations on the door state on the application UI is for better readability. (a) The door is closed; (b) The door is opened, and one person has entered.

The use cases for crowding detection and loitering detection are similar. They are also based on human detections and the camera-related room information pre-stored in the database.

### B. Fire

In this example, a fire scene has been simulated in the virtual hospital's operation room, where one camera and a fire sensor are mounted (Figure 5(a)). The camera view visualised in Smart Client is shown in Figure 5(b-c). Figure 5(b) shows the view before the fire is ignited. When the fire starts, the virtual fire sensor sends the signal to the device driver of XProtect (see Section B) that triggers an alarm, as shown in Figure 5(c). The same simulation has also successfully been used to validate video-based fire detection (See Section 2.3).

---

[11] https://www.boost.org/
[12] https://restfulapi.net/

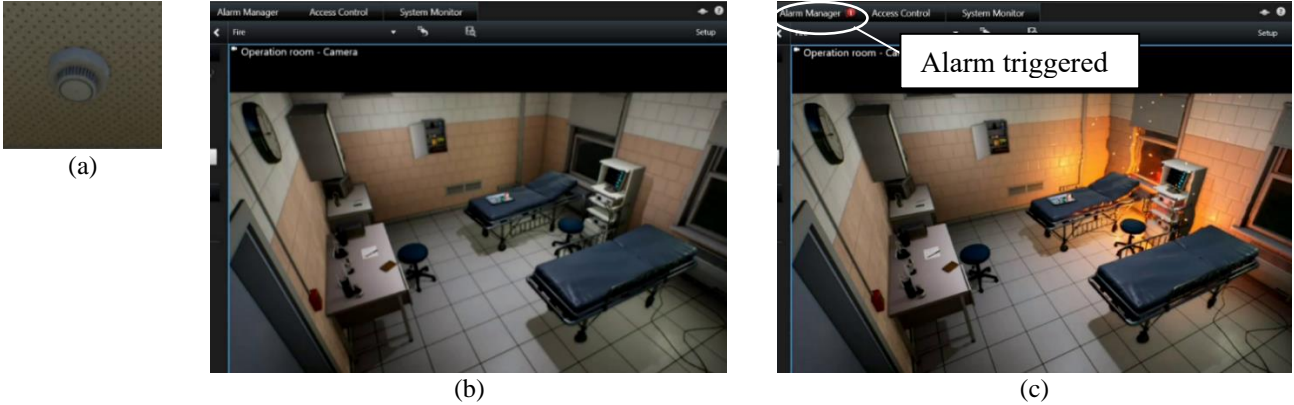

Figure 5. Simulation of fire in a hospital operation room as visualised in Smart Client. (a) Simulated smoke/fire sensor; (a) No fire; (b) Fire is ignited, and the smoke/fire sensor triggers an alarm in the VMS.

Due to similar requirements, the same workflow can be applied for studies on other hazardous incidents, such as flood, explosion and power supply attack. Figure 6 shows a simulation example where the power of the whole building is cut.

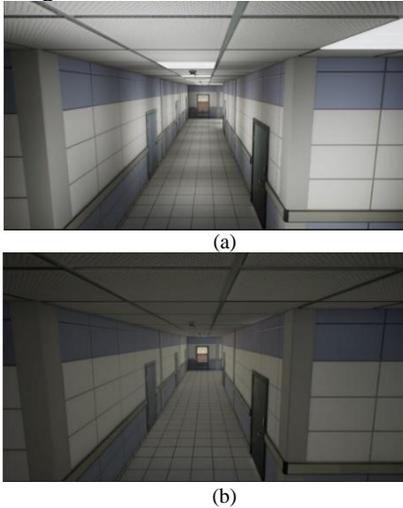

Figure 6. Power supply attack simulation. (a) Normal state; (b) Power is cut.

### C. Weapon Detection

Like person detection, weapon detection using surveillance cameras is also based on visual object detections (Section B. However, three main challenges are often encountered, i.e., the small sizes of many weapons, the weapon diversity and the reflections of the shinning parts [6], and a good dataset needs to cover different situations. Simulated weapons in the virtual world can effectively extend the training and validation dataset.

We have simulated persons carrying different types of weapons, including shotguns, handguns, knives and blunt objects, difficult to provide in real life.

Figure 7 shows four examples of simulated weapons carried by a person, i.e., rifles, pistols, machetes and axes, from varying angles. One character carrying a particular weapon in Unreal can generate images from various angles and with different reflection effects, boosting the training and validation datasets.

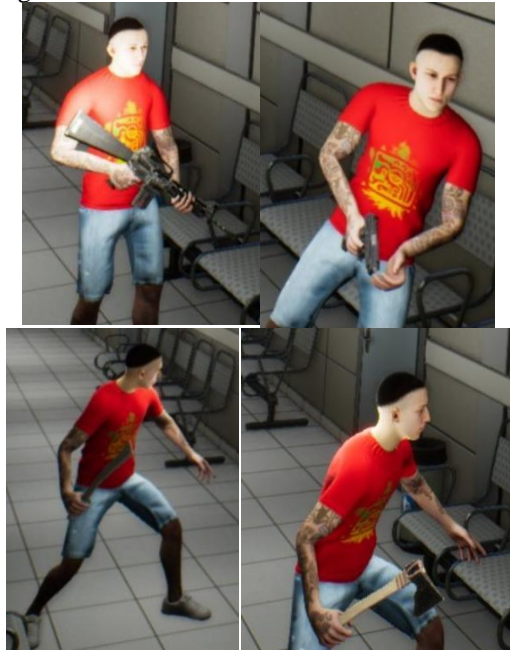

Figure 7. Four examples of simulated weapon detections. (a) Rifle (b) Pistol (c) Machete (d) ax. Credit for weapons: sketchfab.com; license: Attribution 4.0 International ((a) Mateusz Woliński, (b) 3DMaesen, (c) r.m.kowalewski and (d) ALBankrash).

### D. Suggestions for Future Work

The video analysis in our work so far has only focused on individual cameras. It is interesting to extend the study to multi-camera and multi-room person re-identification and involve the building ontologies in the security study. When the virtual world is expanded, one should avoid unnecessary RAM and video memory consumption. Particle systems, such as water, fire and grass, are sources-demanding, and the usage of those should be minimised. A virtual camera should only send video streams when it is connected to an external client application.

For detecting certain threat types, such as tailgating, the camera's location and field of view are crucial. The flexibility of a virtual building allows easy adjustment of cameras. Similar approaches can be applied to studies of camera installation optimisation in a building.

The VMS used in the current set-up is the Milestone XProtect®. However, the implementation of Unreal actors in the virtual building and HTTP/Websocket communication is entirely VMS platform agnostic. With free access to this project's source code, other researchers or engineers can port the solution to other VMS with relatively little effort by wrapping the HTTP/Websocket clients in proper device drives or plugins. The implementation of virtual cameras and communication between the virtual world and external applications is also applicable for other simulation environments such as traffic optimisation.

## V. CONCLUSION

This paper introduces a unifying threat monitoring system with cameras, digital sensors and location information integrated into a well-known VMS, the Milestone XProtect®. Due to the restrictions in data collections and the system's complexity, a virtual smart building has been created. It proves effective in enhancing the datasets for deep learning solution development. A repeatable and flexible test system is provided for complex threat types and for testing hazardous incidents. The software solutions are highly modulated and thereby are ready for reuse in other similar projects. Even though a virtual environment will not replace a real-life physical environment in physical threat monitoring studies, it has significantly facilitated the development and validation process in this area.